\documentclass[letterpaper, 10 pt, conference]{ieeeconf}  
\usepackage{epsfig}
\usepackage{graphicx}
\usepackage{subcaption}

\IEEEoverridecommandlockouts                              

\title{\LARGE \bf
Event-LSTM: An Unsupervised and Asynchronous Learning-based Representation for Event-based Data
}

\author{Lakshmi Annamalai and Vignesh Ramanathan and Chetan Singh Thakur 
}

\begin{document}

\maketitle
\thispagestyle{empty}
\pagestyle{empty}

\begin{abstract}

Event cameras are activity-driven bio-inspired vision sensors, thereby resulting in advantages such as sparsity, high temporal resolution, low latency, and power consumption. Given the different sensing modality of event camera and high-quality of conventional vision paradigm, event processing is predominantly solved by transforming the sparse and asynchronous events into $2D$ grid and subsequently applying standard vision pipelines. Despite the promising results displayed by supervised learning approaches in $2D$ grid generation, these approaches treat the task in supervised manner. Labeled task specific ground truth event data is challenging to acquire. To overcome this limitation, we propose \textit{Event-LSTM}, an unsupervised Auto-Encoder architecture made up of LSTM layers as a promising alternative to learn $2D$ grid representation from event sequence. Compared to competing supervised approaches, ours is a \textit{task-agnostic} approach ideally suited for the event domain, where task specific labeled data is scarce. We also tailor the proposed solution to exploit asynchronous nature of event stream, which gives it desirable charateristics such as speed invariant and energy-efficient $2D$ grid generation. Besides, we also push state-of-the-art event de-noising forward by introducing memory into the de-noising process. Evaluations on activity recognition and gesture recognition demonstrate that our approach yields improvement over state-of-the-art approaches, while providing the flexibilty to learn from unlabelled data.

\end{abstract}

\section{INTRODUCTION}

Event-based cameras, also known as silicon retinas, are a novel type of biologically inspired sensors which encode per-pixel scene dynamics asynchronously with microsecond resolution in the form of a stream of events. Events are triggered as and when the logarithmic change of image intensity $I(x,y,t)$ exceeds a defined threshold. Remarkably, the asynchronous and sparse nature of these event cameras makes the complex vision pipeline computationally more efficient. Other key advantages of event camera are: high dynamic range (140 dB vs. 60 dB of standard cameras) and low power requirements \cite{Delbruck:motor} \cite{Son:dynamic} \cite{Brandli:latency} \cite{Brandli:latency} \cite{Posch:QVGA}, which makes it a suitable choice for resource-constrained environments. However, one of the most challenging aspects of working with event cameras is the lack of frames, thus prompting the pioneers in this field to equip vision algorithms with a paradigm shift that allows efficient extraction of meaningful information from the space-time event data without sacrificing the sparsity and temporal resolution of the event data. 

Inspired by the benchmark set by traditional vision and deep learning approaches \cite{LeCun:Handwritten} \cite{Srivastava:Dropout} \cite{Bengio:Representation} \cite{Krizhevsky:ImageNet} \cite{Shelhamer:Fully} \cite{Xie:Holistically}, one of the predominant area of research in event data focusses on how to aggregate the information conveyed by individual Spatio-temporal events onto a $2D$ grid representation, which enables its compatibility with the tools available from conventional vision. While interest in converting events into $2D$ grid by hand-crafted data transformations is growing, only very few approaches have looked into the more complex solutions that data-driven deep learning methods can provide. The recent supervised deep-learning works proposed are \cite{Daniel:End}, and \cite{Cannici:Matrix}. However, not every application has enough volume of labeled data to quench the data-hunger thirst of supervised deep learning technologies, limiting the design of supervised deep networks to approximate complex functions. Though supervised extraction of features is robust, the key challenge lies in extracting domain-specific features for each task. Convergence is guaranteed only if sufficiently labeled data is available; otherwise, the networks are more prone to get stuck in local minima, particularly for event data that is highly non-linear. Hence, the palpable advantage could be attained if the problem could be formulated to avoid over-fitting the training data caused by the dearth of learning parameters.

The main contribution of this paper is a generic, deep learning-based task-independent architecture (\textit{Event-LSTM}) for integrating raw events into $2D$ grid features. We achieve task independence by operating the popular architecture Long Short-Term Memory (LSTM) network in an unsupervised setting to learn a mapping from raw events into a \textit{task-unaware} $2D$ grid, which we call as LSTM Time Surface (\textit{LSTM-TS}). The proposed \textit{Event-LSTM} puts forth unsupervised event data $2D$ grid generation as an alternative to data-hungry supervised learning approaches while still allowing us to utilize the data-driven learning capability of deep neural networks. It can also pave the way to enable modeling of complex structures by stacking multiple independently trainable feature extraction layers, in addition to eliminating the need for large quantities of labeled data for each task at hand. The proposed architecture is an effective replacement for hand-crafted features, which learn $2D$ grid in an unsupervised fashion from raw event data not customized to the task at hand. The learned model could also be utilized to initialize the supervised architectures.

Driven by the need for efficient speed invariant feature extraction and to take advantage of the asynchronous sensing principle of event cameras, we propose an asynchronous method of windowing of event sequence. A window dependant on time can result in motion blur or under-representation of the environment. As the event sensing modality depends on the dynamics of the scene, the best windowing strategy will be the one that is intrinsically dependant on the number of events. This fully leverages the event camera's benefits to sense speed independent motions of the object, resulting in substantial scene dependant computational power saving. Despite the fact that the asynchronous windowing has been explored in literature previously \cite{Vasco:Harris} \cite{Calabrese:DHP19} \cite{Alzugaray:Asynchronous} \cite{Manderscheid:Invariant}, deep learning-based asynchronous $2D$ grid generation remains unexplored.

The event camera's sensitivity to temporal noise and junction leakage currents results in events generated under constant illumination with no activity in the scene, referred to as BA (Background Activity) noise. To remove BA noise, spatiotemporal filters \cite{Khodamoradi:Spatiotemporal} \cite{Feng:Density} \cite{Wu:Denoising} have been suggested in the literature, which filters out noise based on its correlation with other events in the Spatio-temporal neighborhood. These filters, on the other hand, are overly sensitive to changes in event patterns. In this work, we propose an effective formulation that enables us to filter the noise based on decaying memory of past events in the spatiotemporal neighborhood, thus making it more noise resistant.

Summing up, we make three contributions: \textit{1) To address the scarcity of labeled event data, a generic unsupervised LSTM-based deep learning architecture named as \textit{Event-LSTM} has been proposed to extract meaningful information from raw events}, \textit{2) Motion invariant and energy-efficient feature extraction by rendering $2D$ grid representation within an event-driven window.} and \textit{3) A robust de-noising algorithm designed with the memory of past spatiotemporal events.}

\begin{figure*}
  \centering
  
    \includegraphics[width=\textwidth,height=4in,keepaspectratio=true]{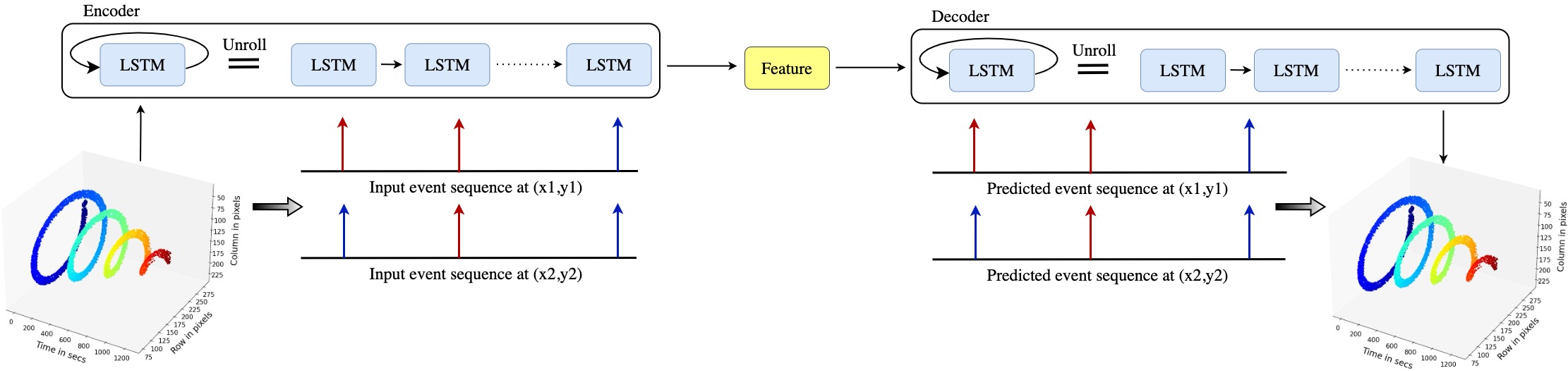}

  \caption{Architecture of the proposed \textit{Event-LSTM} during training phase. The event sequence at each pixel is grouped in synchronous/asynchronous mode and fed to the encoder of LSTM one by one. The encoder and decoder are trained to learn a good feature at each pixel and to reconstruct the corresponding event sequence, respectively.}
  \label{fig:framework}
\end{figure*}

\section{RELATED WORK}

This section involves a brief review of the event-based feature extraction methods focusing on $2D$ grid representations of event data, highlighting the advantages of our proposed method. A detailed review is provided at \cite{Guillermo:Survey} \cite{LakshmiAandAnirbanChakraborty}. The most commonly used framework for an asynchronous event by event processing is a promising research field known as spiking neural networks (SNN) \cite{Arnon:gesture} \cite{Haeng:spiking} \cite{Perez:Mapping} \cite{Dongsung:Gradient} \cite{ThakurCSandJamalMolin}. However, there is a lack of standard training procedures due to their non-differentiability. The alternative line of research is to perform pre-processing on events to convert them into formats compatible with conventional image-based vision architectures. Combined with a high signal-to-noise ratio compared to raw events, the traditional, proven vision solutions have resulted in high accuracy. However, this involves coming up with a better and optimal choice of converting raw asynchronous event stream into conventional vision algorithm compatible structures. This effort has led to different initial representation stages such as event image \cite{Maqueda:steering} \cite{Rebecq:EVO} \cite{Gehrig:EKLT} \cite{Liu:optical} \cite{Kogler:stereovision} \cite{Maqueda:FlowNet}, Time Surface ($TS$), voxel grid \cite{Alex:Unsupervised} \cite{Henri:Bringing} and motion compensated event image \cite{Gallego:unifying} \cite{Gallego:angular}. 

Researchers came up with a popular $2D$ representation called Time Surface (\textit{TS}), which stores a time value at each pixel. This results in encoding motion history at each pixel, thus making it sensitive to the motion of the edges. This gives \textit{TS} an edge over other $2D$ representations, especially for the vision tasks that involve motion analysis. Recently, $TS$ was introduced in \cite{Lagorce:Hots} which proposed a hierarchial representation learned at each layer based on $TS$ prototypes. The major disadvantages are its latency to learn the projection at each layer and its high susceptibility to noise as they were utilizing only the recent event, which limits it from capturing the motion of the history. Hence, \cite{Amos:HATS} proposed a fixed-length memory-based $TS$ and achieved state-of-the-art results in object recognition by extracting histograms from $TS$ as features, known as Histogram of Averaged Time Surfaces (HATS) followed by Support Vector Machine (SVM) for classification. The advantage is that it could be updated asynchronously provided a good compute facility is available. However, the event feature generation procedure proposed so far are hand-crafted.

Coupling deep learning methods with event data will provide us the flexibility to take advantage of event data as well as the learning algorithms. In \cite{Daniel:End}, authors proposed a Multi-Layer Perceptron (MLP) architecture towards converting event data to $2D$ grid representation. As the architecture is completely defined by differentiable operators, it makes the process completely learnable. The information at each pixel is formed by accumulating the MLP feature generated from the events that occur at the same spatial location. The major disadvantage of this method is that the accumulation of events does not depend on the sequence of occurrence of events, thus inhibiting the network from learning the memory embedded in the events. This has been overcome by \cite{Daniel:Phased} and \cite{Cannici:Matrix} by leveraging the memory property of LSTM. \cite{Daniel:Phased} proposed Phased LSTM, which does not result in explicit formation of $2D$ grid representation \cite{Yin:Graph} \cite{Qinyi:recognition}. This prevents us from leveraging the spatial structure of the data, which could be learned using CNN architectures, thus restricting it to simple applications \cite{Marco:Attention}. Moreover, it also introduced huge latency as the events have to pass through the network sequentially. Hence, \cite{Cannici:Matrix} explored the utility of LSTM as $M\times{N}$ (pixel grid size) matrix of cells to learn the mapping from raw event sequence to a dense $2D$ grid. LSTM cells process the sequence of events at each pixel location, condensing it into a single output vector that populates the $2D$ grid. They have demonstrated good accuracy in object recognition and optical flow estimation by replacing the corresponding architecture's input representation with the MatrixLSTM layer. To make the feature learning process translation invariant, they have also experimented with parameter sharing across LSTM cells.

The work proposed in this paper is related to \cite{Cannici:Matrix}. We are interested in proposing a solution for situations where the process of collecting labeled data is very costly. This highlighted the need to build an unsupervised deep learning solution to learn the best mapping from raw event data to $2D$ grid representation, which is task-independent unlike \cite{Cannici:Matrix}. 

\section{PROPOSED SOLUTION}

This section formalizes the methodology to convert raw events into $2D$ grid representation (Fig. \ref{fig:framework}). An event camera with pixel grid size $M\times{N}$ results in an asynchronous stream of events only on those locations where the scene dynamics undergo a change. The events are tuples with the format $e_i=\left(x_i,y_i,t_i,p_i\right)$, where $\left(x_i,y_i\right)$ represents the pixel location of the $i^{th}$ event, $p_i\in\{+1,-1\}$ known as ON and OFF events represents the polarity of the event and $t_i\geq{0}$ indicates the timestamp of the event with microseconds resolution. The time of occurrence of an event is a continuous parameter, and hence the number of events with the same timestamp will be extremely limited to get processed. This mandates the need to accumulate the events over a period $\delta{T}$ to process effectively. This representation of events will more robustly describe the dynamics of the scene.

\subsection{Denoising: SpatioTemporal Filter with Memory}
Event camera comes with an intrinsic type of noise called Background Activity  (BA) noise. This refers to the events that occur when there is no significant change in the pixel intensity threshold. This is mainly contributed by thermal noise and junction leakage currents \cite{Lichesteiner:Latency} \cite{Tian:Analysis} \cite{Fowler:Analysis}. Distinctive characteristic of BA noise is the lack of correlation of these events with other events in the spatiotemporal neighborhood. Utilising this property, \cite{Khodamoradi:Spatiotemporal} \cite{Feng:Density} has proposed variants of spatiotemporal correlation filter which filters out events based on the following formulation: It declares an event $e(p,x,y,t)$ as signal if $|t-t_{mn}|\leq{\delta{T}}$ for all $|m-x|\leq\delta{x},|n-y|\leq\delta{y}$, where $t_{mn}$ is the timestamp of the most recent event in that window. The problem with this filter is that filtering happens by considering only the most recent event in the given neighboring window. As a consequence, noisy events can not always be filtered. To mitigate this effect, in this paper, we propose a filter that takes into account the evidence provided from all the past events within the given spatiotemporal window. Only when $\sum_{mn}{\exp\left({t}-t_{mn}\right)}$ exceeds a pre-defined threshold, where $|m-x|\leq\delta{x},|n-y|\leq\delta{y}$, does the filter declares an event $e(p,x,y,t)$ as signal. This injects memory into the filter and ensures that the noisy events are filtered better while the events representing the signal dynamics alone are preserved.

\subsection{Event-LSTM}
The goal here is to learn a mapping $\phi$ from a set of raw events sequence represented as $E=\{e_i\}$ to a $2D$ grid representation \textit{LSTM-TS} of size $M\times{N}$. The mapping should be learned in an unsupervised manner, based solely on the data and independent of the task at hand to account for the situations where the amount of labeled data available for each task is very limited in relation to the number of parameters to be learned. Hence, in this work, we propose an LSTM-based autoencoder structure known as \textit{Event-LSTM}, which learns an encoder and decoder function $\phi$ and $\psi$ respectively to predict feature and reconstruct the input sequence respectively. To ensure translation invariance, our \textit{Event-LSTM} is comprised of a single autoencoder LSTM across all pixels. During the training phase, the event sequence at each pixel location is processed sequentially. However, during the feature extraction phase, processing can be applied in parallel across all pixels.

\begin{figure}
  \centering
    \includegraphics[width=\linewidth,height=4in,keepaspectratio=true]{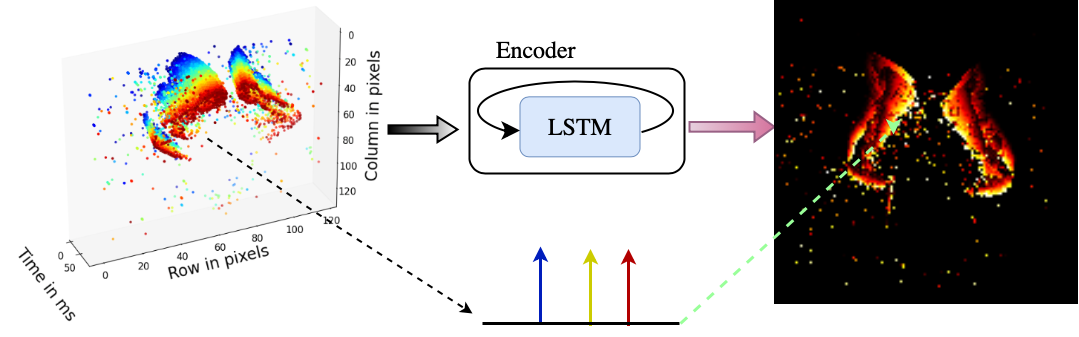}
   
  \caption{Architecture of the proposed \textit{Event-LSTM} during feature extraction phase. The event sequence at each pixel is grouped in synchronous/asynchronous mode and fed to the encoder of LSTM. Only encoder is used during the feature extraction phase. The features extracted at each pixel is populated into a $2D$ grid known as \textit{LSTM-TS}}
  \label{fig:frameworkTest}
\end{figure}

Let $E_{xy}=\left(x_n,y_n,t_n,p_n\right)$ denote the set of events generated at $\left(x_n=x,y_n=y\right)$ with varying lengths $\vert{E_{xy}}\vert$ depending on the dynamic activity level at that specific pixel. Each event sequnce $E_{xy}$ is represented by a sequence of time stamps $t_n$ occurring at that specific location $\left(x_n=x,y_n=y\right)$ ($T_{xy}=\{t_n|x_n=x,y_n=y\}$). LSTM auto-encoder is fed with these time stamp sequences $T_{xy}$. \textit{Event-LSTM} with its internal memory learns an encoder ($\phi(T_{xy}$)) transformation that estimates a meaningful feature \textit{LSTM-TS}$_{xy}$ from the input $T_{xy}$ sequence and a decoder transformation ($\psi\left(\phi\left(T_{xy}\right)\right)$) that tries to recreate the input sequence by minimising the error $\vert{T_{xy}-\widehat{T}_{xy}}\vert$. The $M\times{N}$ \textit{LSTM-TS} is realized by populating a $2D$ grid with the features learned by encoder at each $(x,y)$ location \ref{fig:frameworkTest}. This preserves the sparsity of the event data as computation of features occurs only at locations $\left(x,y\right)$ where the sequence $T_{xy}$ is non-empty.

\subsection{Windowing}

The proposed \textit{Event-LSTM} operates on temporal windows. The input event sequence  $E_{xy}$ is split into smaller sequences  $E^l_{xy}$, resulting in independent  time stamp sequence $T^l_{xy}$ for each window $\delta$. This makes sure that \textit{Event-LSTM} does not have to deal with very long sequences. Although LSTM is better at handling vanishing gradient problems effectively better than RNN, splitting long sequences into smaller sequences ensures that \textit{Event-LSTM} can effectively retain local time information. As described in the sections below, we have experimented with two different types of windowing: synchronous \textit{LSTM-TS-S} and asynchronous \textit{LSTM-TS-A}.

\subsubsection{Synchronous Windowing}
In synchronous binning, the smaller event sequences are formed by dividing, the longer sequence into $L$ bins with equivalent time intervals $\delta{t}$, such that $E^l_{xy}$ consists of all the events whose time of occurrence falls within the interval $\delta{t}$. This will result in $L=\frac{T}{\delta{t}}$ sequences where $T$ is the total time duration of the parent sequence.

\subsubsection{Asynchronous Windowing}
We also propose a type of temporal windowing known as asynchronous temporal windowing, in which the smaller sequences are formed such that the number of total events $\vert{E^l}\vert$ over all the pixels remains constant across $L$ sequences. This will result in $L=\frac{\vert{E}\vert}{\vert{E^l}\vert}$ sequences, with $\vert{E}\vert$ representing the total number of events in the parent sequence and $\vert{E^l}\vert$ representing the number of events in the individual sequence. This form of binning's significant benefits are that it makes the learned features motion invariant and saves computational resources while the scene is idle. This claim has been substantiated with empirical experiments in Section \ref{section:exp}.

\section{EXPERIMENTS AND RESULTS}

\label{section:exp}
The experimental section is devoted to support our claims that the propsoed approach i) improvises the de-noising algorithm by injecting memory into it, ii) infers a suitable asynchronous mapping from event raw domain to $2D$ grid representation and iii) outperforms state-of-the-art unsupervised hand crafted $2D$ grid feature extraction methods in the motion analytics tasks such as activity recognition and gesture recognition.

\subsection{Denoising Error Analysis}
This section shows the influence of inculcating memory into event de-noising approach.

\begin{figure}
  \centering
  \begin{subfigure}{0.48\linewidth}
    \includegraphics[width=\textwidth]{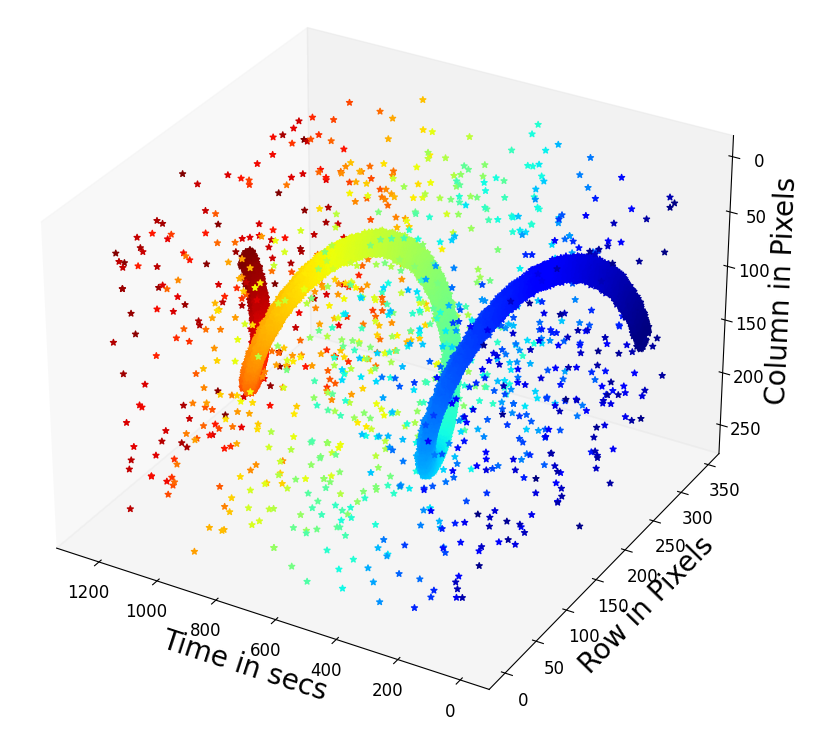}
    \caption{Input}
  \end{subfigure}
  \begin{subfigure}{0.48\linewidth}
    \includegraphics[width=\textwidth]{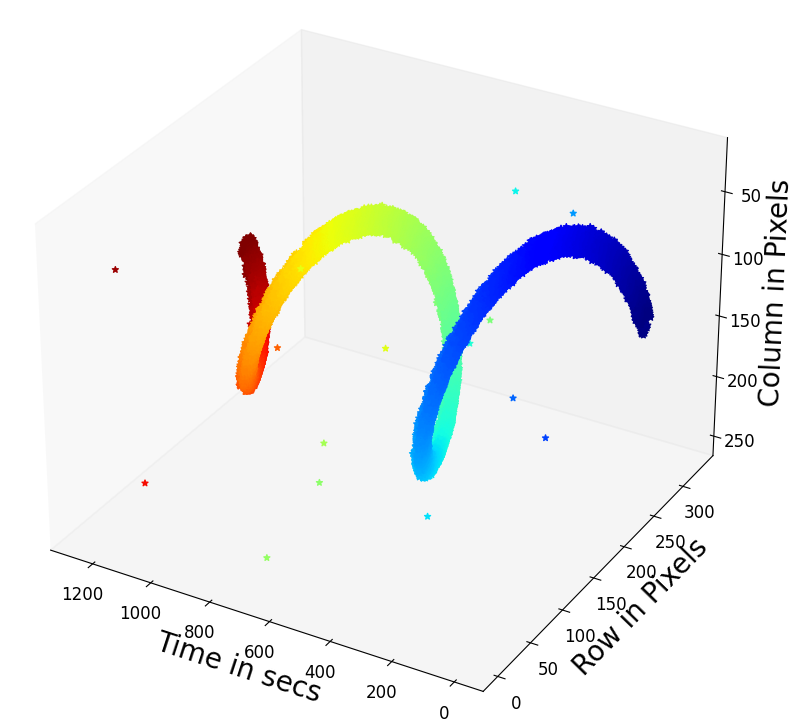}
    \caption{Output}
  \end{subfigure}
  \caption{Input and De-noised events with shot noise added}
  \label{fig:noise_1}
\end{figure}

\begin{figure}
  \centering
  \begin{subfigure}{0.48\linewidth}
    \includegraphics[width=\textwidth,height=1.5in]{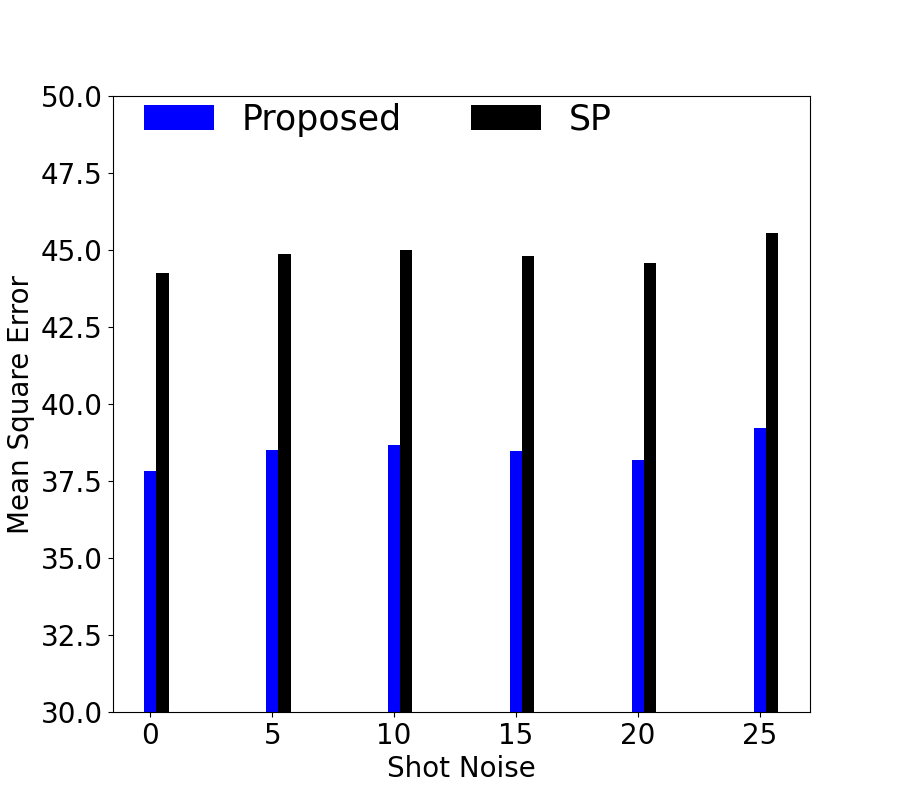}
    \caption{Signal with noise}
  \end{subfigure}
  \begin{subfigure}{0.48\linewidth}
    \includegraphics[width=\textwidth,height=1.5in]{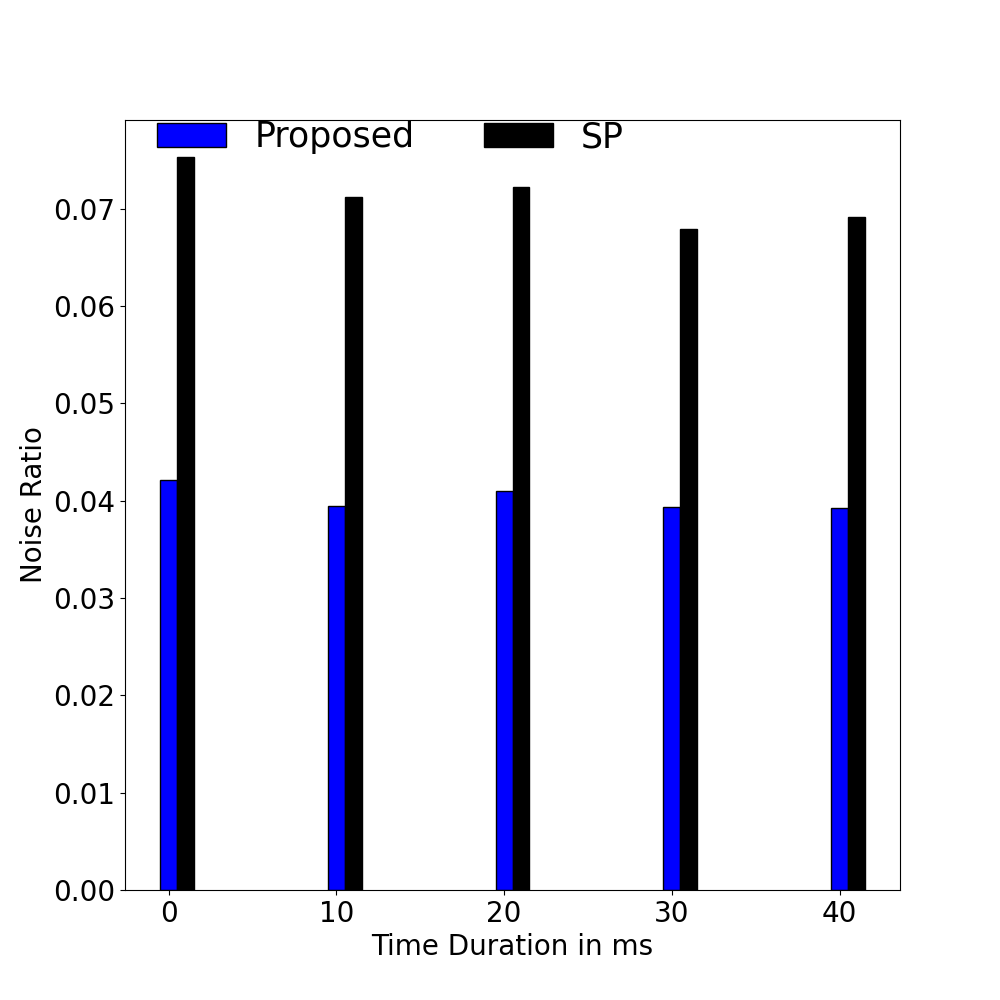}
    \caption{Only noise}
  \end{subfigure}
  \caption{Comparison of the proposed algorithm with Spatio-Temporal denoise filter in the presence of signal with shot noise simulated at different frequencies and in the presence of noise alone (from initial part of DAVIS recordings)}
  \label{fig:noise_2}
\end{figure}

To see how effective the filter is at eliminating the noise while retaining the signal, we ran the following experiment: We chose a wide range of shot noise frequencies, added it to the events simulated (using the event data simulator proposed in \cite{Yuhuang:V2E}) and evaluated the result of de-noising algorithm. Fig. \ref{fig:noise_1} shows visualization of one such sample, whereas Fig. \ref{fig:noise_2} shows the plot of comparison  of the proposed method with \cite{Khodamoradi:Spatiotemporal} in terms of $MSE={\vert{E-\widehat{E}}\vert}^2$ vs. shot noise frequency, where $E$ and $\widehat{E}$ are signal and de-noised events. The proposed method displays lesser $MSE$ and, as a result, higher SNR, thus supporting our claim.

To evaluate the false positive error of the proposed method, we collected the output events of DAVIS sensor in a environment where there was no change in scene dynamics. We repeated this process at different times and used this data to compare the noise eliminating capability of the filter. Fig. \ref{fig:noise_2} shows the comparison of the proposed method with \cite{Khodamoradi:Spatiotemporal} in terms of Noise Ratio ($NR=\frac{N_e^o}{N_e^i}$) estimated at various time instants, where $N_e^o$ and $N_e^i$ are the number of events in the output and input respectively. It is evident that the $NR$ is lower for the proposed algorithm, thus substantiating the better de-noising capability of the same.
 
\subsection{LSTM-TS: Why Asynchronous?}

This section is dedicated to highlight the efficacy and necessity of asynchronous temporal binning of event data. Evaluation is performed on simulated data. Asynchronous binning was set to $\vert{E^l}\vert=1100$ events, and synchronous binning was set to $\delta{t}=1200ms$.

\subsubsection{Speed Invariant Classification}

\begin{figure}
  \begin{subfigure}[b]{0.48\linewidth}
    \includegraphics[width=\textwidth]{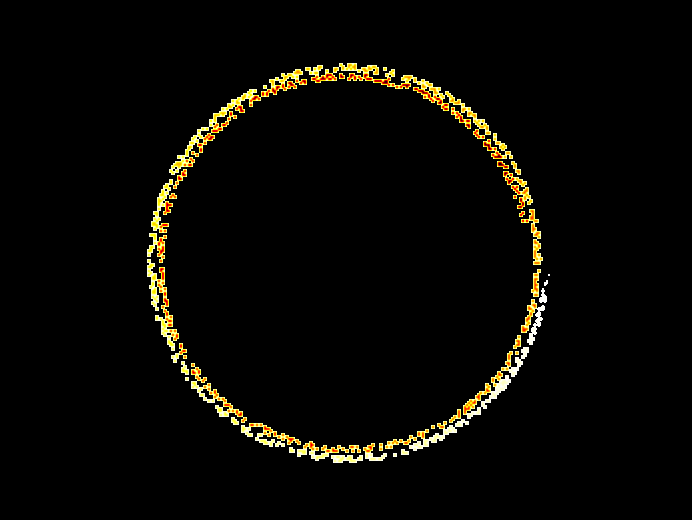}
    \caption{\textit{LSTM-TS-A} (Freq: $1.6$ Hz)}
    \label{fig:2}
  \end{subfigure}
  \begin{subfigure}[b]{0.48\linewidth}
    \includegraphics[width=\textwidth]{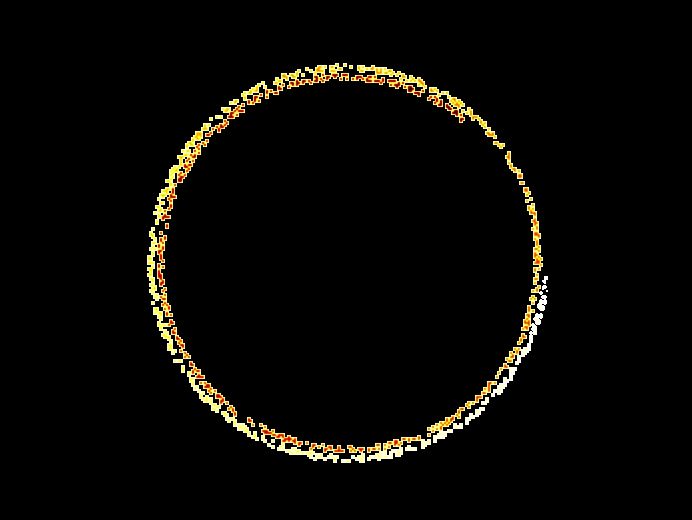}
    \caption{\textit{LSTM-TS-A} (Freq: $3.2$ Hz)}
    \label{fig:1}
  \end{subfigure}
  
  \begin{subfigure}[b]{0.48\linewidth}
    \includegraphics[width=\textwidth]{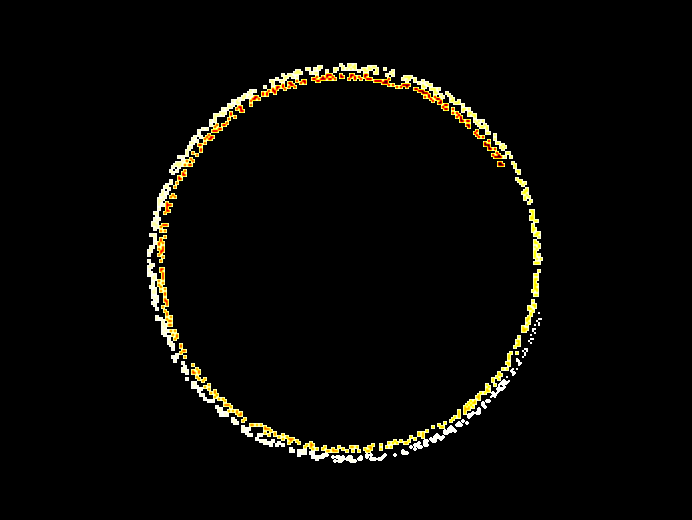}
    \caption{\textit{LSTM-TS-S} (Freq: $1.6$ Hz)}
    \label{fig:1}
  \end{subfigure}
  \begin{subfigure}[b]{0.48\linewidth}
    \includegraphics[width=\textwidth]{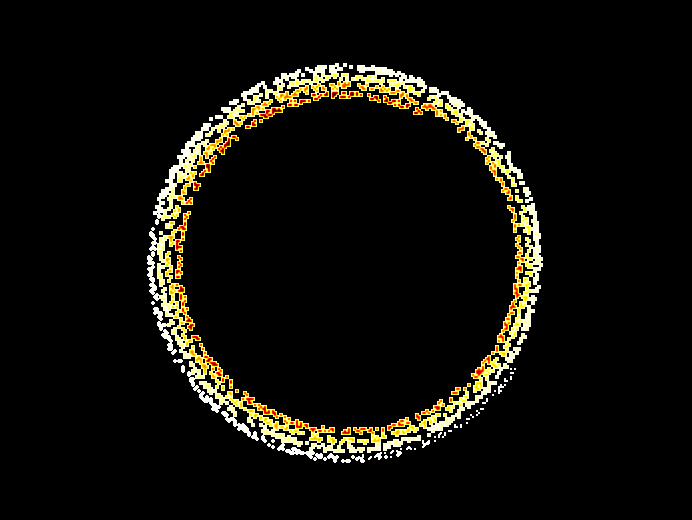}
    \caption{\textit{LSTM-TS-S} (Freq: $3.2$ Hz)}
    \label{fig:2}
  \end{subfigure}
  \caption{Top row: \textit{LSTM-TS-A} ($\vert{E^l}\vert=1100$ events). Bottom row: \textit{LSTM-TS-S} ($\delta{t}=1200 ms$). It could be seen that the features of \textit{Event-LSTM-A} are invariant to the frequency of the rotating dot, hence making it speed invariant.}
  \label{fig:lstm_2}
\end{figure}

To capture the intra-class similarities, the features extracted should be speed invariant. To demonstrate that the extracted \textit{LSTM-TS} in asynchronous mode captures the pattern of motion while remaining invariant to the speed of motion, we simulated a moving dot in the plain background with two different frequencies, one with $2$ cycles (\textit{dotTwo}) and other with $4$ cycles (\textit{dotFour}) in a duration of $1200ms$. With the exception of the speed of motion, these two scenarios are identical. As a consequence, a good feature should result in similar structure for \textit{dotTwo} and \textit{dotFour}. Fig. \ref{fig:lstm_2} shows the extracted \textit{LSTM-TS} in synchronous and asynchronous mode. In the case of asynchronous binning, the feature generated for \textit{dotTwo} resembles that of \textit{dotFour} whereas, in the case of synchronous binning, they create completely different patterns. Note that our asynchronous binning approach results in speed invariant feature extraction necessary to cope up with intra class variations.

\subsubsection{Energy Saving Feature Extraction}

\begin{figure}
  \centering
  \begin{subfigure}{0.48\linewidth}
    \includegraphics[width=\textwidth]{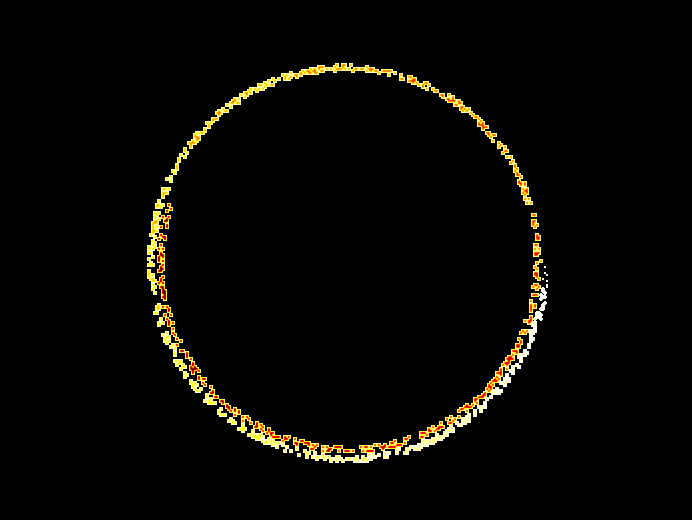}
    \caption{\textit{LSTM-TS-A}}
  \end{subfigure}
  \begin{subfigure}{0.48\linewidth}
    \includegraphics[width=\textwidth]{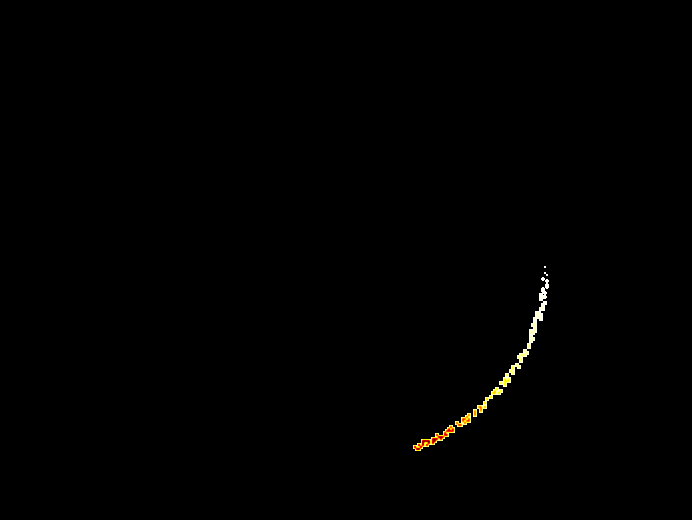}
    \caption{\textit{LSTM-TS-S}}
  \end{subfigure}
  \caption{Input event sequence was a moving dot with a very low frequency of $0.17$ Hz with three cycles. Right: \textit{LSTM-TS-A} ($\vert{E^l}\vert=1100$ events resulting in $2$ \textit{LSTM-TS-A}). Right: \textit{LSTM-TS-S} ($\delta{t}=1200 ms$). It could be seen that synchronous mode was not able to capture the motion of the moving dot. Approximately $15$ \textit{LSTM-TS-S} feature maps were produced, resulting in $7.5$ times more processing than its asynchronous counterpart.}
  \label{fig:lstm_3}
\end{figure}

An event camera's major advantage is that it generates non-redundant data, which results in lesser processing requirements and therefore consumes lesser power. Vision processing algorithms should be able to deal with this remarkable property of the event camera. Asynchronous binning initiates processing only when a specified number of events is accumulated and this may explain the larger flexibility that we could obtain in terms of energy efficient processing. To illustrate this, we have simulated a moving dot with very low frequency, resulting in a significantly lower number of events being generated in a given period. Results of \textit{LSTM-TS} extracted in synchronous and asynchronous mode are given in Fig. \ref{fig:lstm_3}. For an event count of $\vert{E^l}\vert=1100$, asynchronous mode resulted in two \textit{LSTM-TS} capturing enough information regarding the dot's motion history. In contrast, with the fixed time interval of $\delta{t}=1200ms$ (as mentioned earlier), synchronous mode resulted in $15$ such \textit{LSTM-TS}, with not much information captured in each \textit{LSTM-TS}. This shows that asynchronous setting is indeed computationally efficient resulting in non-redundant \textit{LSTM-TS} while still extracting richer information from the raw event sequence.

\subsection{LSTM-TS on Applications: Activity Recognition and Gesture Recognition}

We tested the proposed solution on a prevalent vision tasks known as activity recognition \cite{Michael:action} and gesture recognition on the datasets provided at \cite{Miao:action} and \cite{Arnon:lowpower} respectively. \cite{Michael:action} dataset is a collection of recordings from an empty office captured with DAVSI346. The dataset comes with 15 subjects acting $12$ different actions, with each action lasting $5s$. There is significant variation in the motion speed among the subjects, as presented in the analysis of the paper. The gesture recognition dataset \cite{Arnon:lowpower} comprises $11$ hand and arm gestures comprising of $1595$ instances. It has been collected from $29$ subjects. As the proposed method primarily targets the scenario where the number of labeled data is minimal, we have used only $319$ instances. In order to prove the task-independent nature of \textit{Event-LSTM}, we trained it on activity recognition data alone. Hyperparameter ($|E^l|$ and $\delta{t}$) search has been carried out empirically. We then select the integration interval as $|E^l|=10k$ events (asynchronous mode) and $\delta{t}=100ms$ (synchronous mode). Input time features are normalized between $0$ and $1$.

\subsubsection{Visualization of LSTM-TS-A}

\begin{figure}
  \centering
   \begin{subfigure}{0.48\linewidth}
    \includegraphics[width=\textwidth]{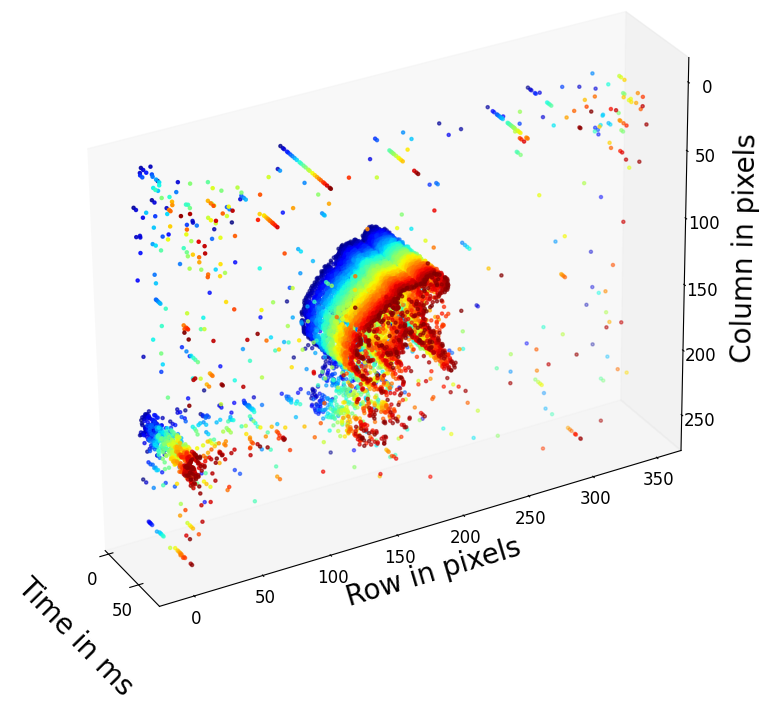}
    \caption{\textit{Event-LSTM} Input}
  \end{subfigure}
  \begin{subfigure}{0.48\linewidth}
    \includegraphics[width=\textwidth]{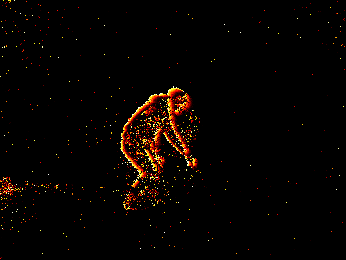}
    \caption{\textit{LSTM-TS-A}}
  \end{subfigure}
  \caption{Visualization of the $3D$ event input and output of \textit{EventLSTM} of Picking Activity. Left: $3D$ input to \textit{Event-LSTM}. Time stamp color coded. Right: \textit{LSTM-TS-A}. Intensity value of features color coded.}
  \label{fig:act_1}
\end{figure}

This section gives visualisation of the $3D$ input and output of \textit{EventLSTM} on a picking activity sample of activity recognition dataset (Fig. \ref{fig:act_1}). 

\subsubsection{Synchronous vs. Asynchronous}
\label{section:act}

The two configurations of \textit{Event-LSTM} are comapared in terms of multi-class metrics such as precision, recall and F1-Score over a wide range of activities and gestures. Towards this, we start out with providing time stamp of a limited number of recordings of \cite{Miao:action} as input to \textit{Event-LSTM}. The model was frozen after having been trained with unlabelled activity data. A limited number of recordings has been used to extract \textit{LSTM-TS} for each activity and gesture. We use \textit{MobileNet} as primary feature extractor of \textit{LSTM-TS} pre-trained on ImageNet. This was followed by the popular classifier referred as Support Vector Machine (SVM). We remark that the only network which requires labeled data is SVM. Since SVM is a shallow classifier, the amount of labeled data needed is substantially lesser than that demanded by deep networks. Results of the two variants are reported in Table. \ref{table:act_1} and \ref{table:gest_1}. Asynchronous setting performs consistently better on a wide range of activities and gestures.

\begin{table}[]
\centering
\begin{tabular}{c|cc|cc|cc}
\hline
\textbf{Activity} &\multicolumn{2}{c|}{\textbf{Precision}} &\multicolumn{2}{c|}{\textbf{Recall}} & \multicolumn{2}{c}{\textbf{F1-Score}} \\ \hline
  & Async & Sync &Async &Sync &Async &Sync \\ \hline
\textbf{Armcross} & 0.67     & 0.67          & 1.00      &0.57      & 0.80    &0.62          \\
\textbf{Picking}  & 0.92     & 0.75          & 0.85      &1.00      & 0.88    &0.86          \\
\textbf{Falling}  & 1.00     & 1.00          & 1.00      &0.90      & 1.00    &0.95          \\
\textbf{Waving}   & 1.00     & 0.75          & 1.00      &0.60      & 1.00    &0.67          \\
\textbf{GetUp}    & 1.00     & 0.95          & 0.78      &1.00      & 0.88    &0.86          \\
\textbf{Walking}  & 0.92     & 0.80          & 0.92      &0.95      & 0.92    &0.95          \\
\textbf{Turning}  & 1.00     & 0.93          & 1.00      &0.80      & 1.00    &0.80          \\
\textbf{Sit}      & 1.00     & 0.67          & 1.00      &0.93      & 1.00    &0.93          \\
\textbf{Kicking}  & 0.50     & 1.00         & 1.00       &0.67     & 0.67     &0.67         \\
\textbf{Jumping}  & 0.80     & 0.75         & 1.00       &0.67     & 0.89     &0.80         \\
\textbf{Tying}    & 0.88     & 0.75         & 0.88       &0.60     & 0.88     &0.67         \\
\textbf{Throwing} & 1.00     & 0.75          & 1.00      &0.86      & 1.00    &0.80          \\ \hline
\end{tabular}

\caption{Analysis of \textit{LSTM-TS-A} vs. \textit{LSTM-TS-S} on activity recognition in terms of precision, recall and F1-Score}
\label{table:act_1}
\end{table}

\begin{table}[]
\centering
\begin{tabular}{c|cc|cc|cc}
\hline
\textbf{Gesture} &\multicolumn{2}{c|}{\textbf{Precision}} &\multicolumn{2}{c|}{\textbf{Recall}} & \multicolumn{2}{c}{\textbf{F1-Score}} \\ \hline
  & Async & Sync &Async &Sync &Async &Sync \\ \hline
\textbf{LH\_C} & 0.86       &0.76        & 0.93      &0.96      & 0.89     &0.85         \\
\textbf{FA\_RB}  & 0.92        &0.92       & 0.98       &0.84     & 0.95      &0.89        \\
\textbf{RH\_W}  & 0.93          &0.94     & 0.96       &0.92     & 0.94        &0.92      \\
\textbf{LH\_CC}   & 1.00      &1.00         & 0.1    &0.33        & 0.18     &0.50         \\
\textbf{RH\_C}    & 0.87          &0.80     & 0.81       &0.75     & 0.84     &0.77         \\
\textbf{LH\_W}  & 0.98         &0.93      & 0.94       &0.93     & 0.96         &0.93     \\
\textbf{Clap}      & 0.95          &1.00     & 0.77       &0.83     & 0.85         &0.91     \\
\textbf{Drums}  & 0.91          &0.86     & 0.91       &0.82     & 0.91  &0.84            \\
\textbf{RH\_CC}  & 0.80    &0.79           & 0.88    &0.79        & 0.84 &0.79             \\
\textbf{Guitar}    & 0.91        &0.64       & 0.89       &0.69     & 0.90        &0.67      \\ \hline
\end{tabular}


\caption{Analysis of \textit{LSTM-TS-A} vs \textit{LSTM-TS-S} on Gesture recognition in terms of precision, recall, and F1-Score. RH-Right Hand, FA-Fore Arm, LH-Left Hand, C-Clockwise, CC-CounterClockwise, W-Wave.}
\label{table:gest_1}
\end{table}

\subsubsection{Perfomance w.r.t State-of-the-art}

For further validation of the proposed solution, we demonstrate that \textit{Event-LSTM} is a better-suited solution than state-of-the-art methods in the context of activity and gesture recognition. The proposed method \textit{Event-LSTM} is an unsupervised feature extraction process of estimating a mapping to convert raw event sequence into a $2D$ structure. To keep the comparison fair, we have considered only unsupervised methods of generating a $2D$ grid.  The various state-of-the-art methods used for comparison are Surface of Active Events (SAE: the timestamp of the recent event), SNN (Count of number of firing spikes), EvOn (Number of on events), EvOff (Number of off events) \cite{Maqueda:FlowNet}, EvCount (number of events at each pixel in the given count of total events) \cite{Calabrese:DHP19} and Exp (time stamp weighted with exponential function) \cite{Amos:HATS}. Through an extensive evaluation, we show that using \textit{LSTM-TS-A} improves the performance of multi-class activity classification and gesture recognition over hand-crafted features by a good margin.

\begin{figure}
  \centering
   \begin{subfigure}{0.48\linewidth}
    \includegraphics[width=\linewidth,height=3in, keepaspectratio=true]{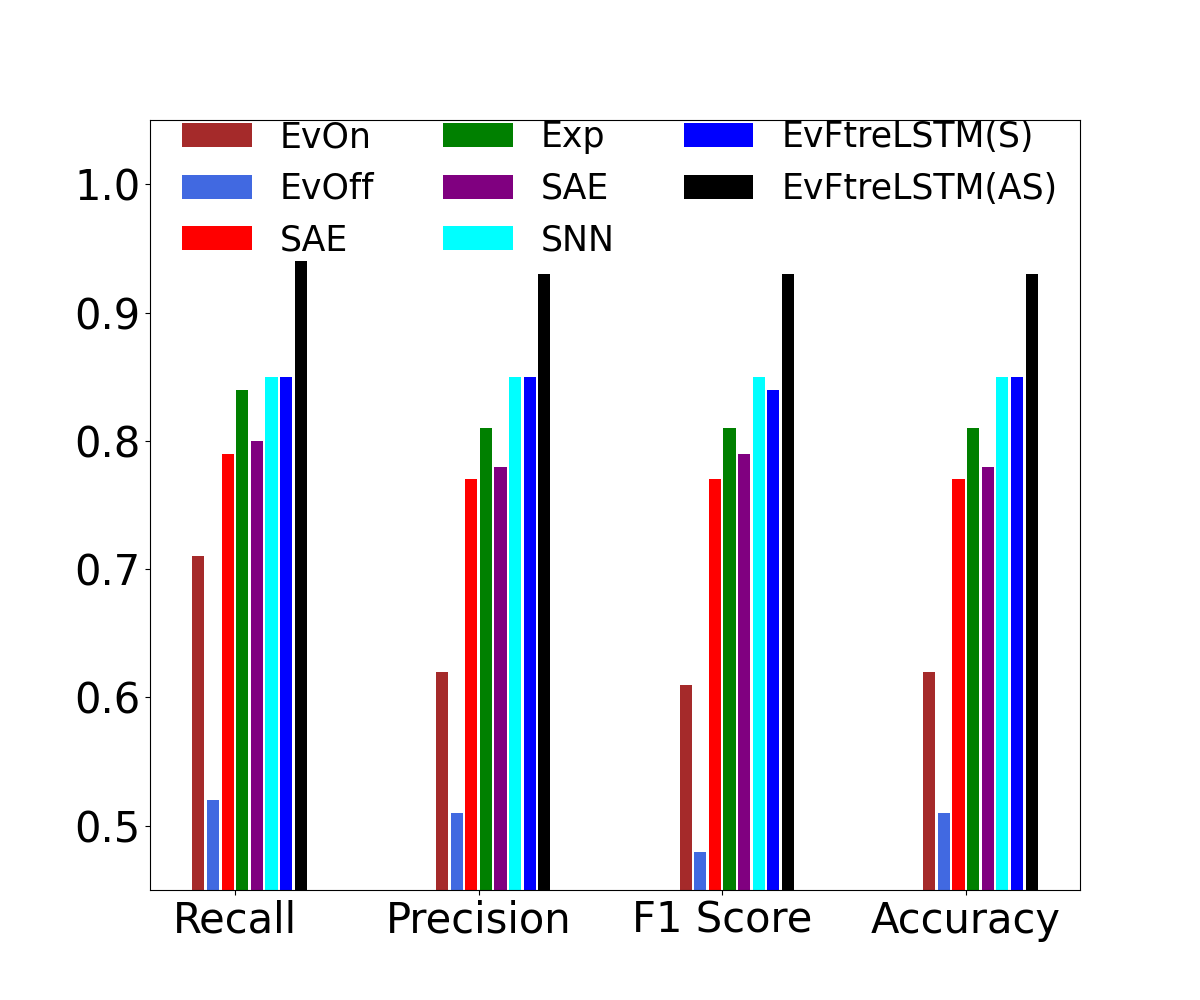}
    \caption{Activity Recognition}
  \end{subfigure}
  \begin{subfigure}{0.48\linewidth}
    \includegraphics[width=\linewidth,height=3in, keepaspectratio=true]{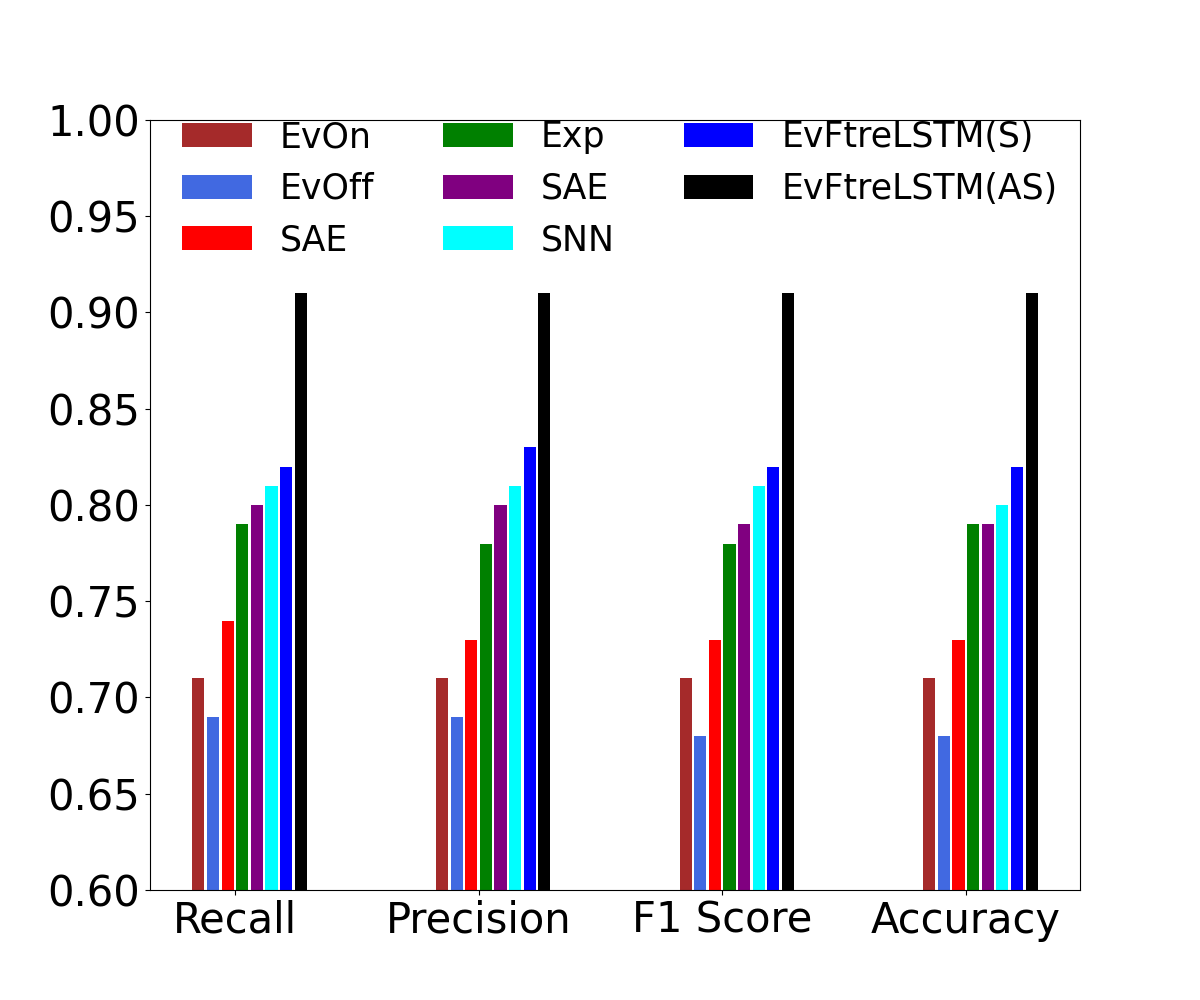}
    \caption{Gesture Recognition}
  \end{subfigure}
  \caption{Comparison of \textit{LSTM-TS} in synchronous and asynchronous mode with State-of-the-art methods on Activity recognition and Gesture recognition. A pre-trained MobileNet architecture (not trained on event activity recognition data) has been used as a feature extractor for the hand-crafted memory surfaces and \textit{LSTM-TS-A} (EvFtreLSTM(AS)) and \textit{LSTM-TS-S} (EvFtreLSTM(S)). Classification is done with SVM, which requires less data for training. Good accuracy shows the utility of the proposed architecture when there is no enough labeled data available to train deep feature extractors.}
  \label{fig:act_2}
\end{figure}

\section{CONCLUSIONS}

In this work, we presented a generic task-independent framework named as \textit{Event-LSTM} to generate $2D$ grid from the sequence of raw events. By modeling the transformation with an unsupervised architecture of LSTM, we have made it a suitable solution to learn features from unlabelled data, especially to mitigate the task specific data-hungerness of supervised deep learning algorithms. Our framework lays out a speed invariant and energy-efficient feature extraction methodology by proposing event dependant windowing. We were also able to boost the noise removal efficiency by injecting memory into the de-noising process. The de-noising capability of the same has been extensively studied under simulated data. Our simulation experiments validates the outstanding property of asynchronous binning. Furthermore, a thorough evaluation of \textit{Event-LSTM} has been carried out on activity recognition, and gesture recognition. Overall, our approach outperforms state-of-the-art unsupervised hand-crafted $2D$ event grid representation, thus enabling us to proceed forward in \textit{data-driven} and \textit{task-unaware} mapping of event data to $2D$ grid representation.




\bibliographystyle{IEEEtran}
\bibliography{references}

\end{document}